\documentclass[conference, 12pt]{IEEEtran}
\usepackage{stmaryrd}
\usepackage{amsfonts}
\pdfoutput=1

\usepackage{graphicx,times,amsmath}

\IEEEoverridecommandlockouts    

\begin{document}
\title{ \LARGE\bf Automatic Skin Lesion Segmentation using Semi-supervised Learning Technique}

\author{
    \IEEEauthorblockN{S. M. Jaisakthi\IEEEauthorrefmark{1}, Aravindan Chandrabose\IEEEauthorrefmark{2}, P. Mirunalini\IEEEauthorrefmark{2}}
    \IEEEauthorblockA{\IEEEauthorrefmark{1}School of Computing Science \& Engineering, VIT University, Vellore, India
    \\{jaisakthi.murugaiyan}@vit.ac.in}
    \IEEEauthorblockA{\IEEEauthorrefmark{2}Department of Computer Science \& Engineering, SSN College of Engineering, Chennai, India}    
    \{miruna,aravindanc\}@ssn.edu.in}

\maketitle

\begin{abstract}
Skin cancer is the most common of all cancers and each year million cases of skin cancer are treated. Treating and curing skin cancer is easy, if it is diagnosed and treated at an  early stage. In this work we propose an automatic technique for skin lesion segmentation in dermoscopic images which helps in classifying the skin cancer types. The proposed method comprises of two major phases (1) preprocessing and (2) segmentation using semi-supervised learning algorithm. In the preprocessing phase noise are removed using  filtering technique and in the segmentation phase skin lesions are  segmented based on clustering technique. K-means clustering algorithm is used to cluster the preprocessed images and skin lesions are filtered from these clusters based on the color feature. Color of the skin lesions are learned from the training images using histograms calculations in RGB color space. The training images were downloaded from the ISIC 2017 challenge website and the experimental results were evaluated using validation and test sets.
  
\end{abstract}

\section{Introduction}
Cancer begins in cells \cite{skin_cancer}, which are the building blocks of tissues. The skin and other organs of the body are made up of tissues. Based on the requirement of the body, normal cells grow and divide to form new cells. Usually new cells take the place of old or damaged cells when they die. This normal process goes wrong, if new cells form even when the body does not need them. This extra cells form a mass of tissue called a growth or tumor. 

Growths on the skin can be benign (not cancer) or malignant (cancer). Benign growths are not as harmful as malignant growths. Benign growths are rarely threats to life and can be removed. Usually benign don't grow, don't intrude into the tissues around them and don't spread to other parts of the body. Malignant growths may be a threat to life, can be removed but sometimes grow back. Melanoma may intrude and damage nearby organs and tissues and may invade other parts of the body where it is very hard to treat.  If detected at an early stage melanomas are curable. 

Generally skin cancer is screened by clinicians through visual examination. During the screening of cancer, the clinicians look for moles and other spots that are different in color from the normal skin. The rule for cancer detection is called as ABCD rule \cite{Lau} which is given by
\begin{itemize}
 	\item A : asymmetry (one half of the mole does not match the other half)
	\item B : border irregularity (edges of the mole are ragged, notched, or blurred)
 	\item C : color (pigmentation of the mole is not uniform, with varying degrees of tan, brown, or black)
 	\item D : diameter of more than ¼ inch (about the size of a pencil eraser)
  	\item E : evolving (the mole is changing over time)
\end{itemize} 
Visual screening of clinicians for skin cancer does not guarantee 100\% detection and sometimes it may lead to potential harm. Potential harm includes unnecessary procedures such as skin biopsy or excision for lesions that do not turn out to be cancer or sometimes the lesions might have missed and not have gone for biopsy, resulting in death. As a result there is a clear requirement for automatic detection system for skin cancer which should be highly efficient and accurate.
In this paper we propose an automatic skin cancer segmentation algorithm which partitions the skin lesions area using semi-supervised learning technique. The proposed method consist of two major steps (1) Preprocessing which removes the artifacts like hair, ink markings and illumination defects. (2) Segmentation of lesions using k-means clustering and color histogram feature.

\section{ Proposed Methodology}
\subsection{Preprocessing}
\subsubsection{Image Scaling}
The size of the images in the training set varies from 1022$\times$ 767 to 6708 $\times$ 4439. To reduce the computational complexity, the images in the training set were resized down to 25\% using bi-linear interpolation method \cite{gonzalez2009digital}, if the height or width of the original image is above 1500. If the width or height of the image is less than 1500 the images size is retained as such.

\subsubsection{Illumination Correction}

To remove uneven illumination of the image caused by sensor faults, non uniform illumination of the scene, or orientation of the objects surface illumination correction technique is applied. The original input image is first converted into Lab color space and then Contrast Limited Adaptive Histogram Equalization (CLAHE ) algorithm \cite{Zuiderveld:1994:CLA:180895.180940} is applied to L channel. The contrast enhanced L channel is merged to form Lab image back and then converted into RGB image.
CLAHE is the upgraded version of Adaptive Histogram Equalization (AHE) which overrides the drawback of standard histogram equalization. In CLAHE algorithm, the input image is divided into small regions and histogram equalization is applied to each of the regions rather than the entire image.

\subsubsection{Hair removal}
Some of the images in the training set contain hair which makes the segmentation process hard.  The presence of hair also makes the feature extraction process for further image analysis demanding. So we have removed the hairs using Frangi vesselness algorithm \cite{frangi1998multiscale}. 
Hairs can be regarded as the tubular structure and hence can be enhanced as a bright line by applying Frangi 2D filter. Most of the 2D filter algorithms are based on 2nd order derivatives. Frangi vesselness filter was introduced by Frangi et.al. \cite{frangi1998multiscale}. The vessel enhancement process searches for geometrical structures which can be considered as tubular. Frangi vesselness filter first determines the second order partial derivative of the image that is Hessian matrix. From the calculated hessian matrix eigenvalue decomposition is applied in order to calculate eigenvalues. The eigenvector corresponding to the smallest eigen value is used to filter hairs in the images.

\subsection{Segmentation}
Segmentation is the process of isolating the diseased area from normal skin based on the homogeneity of the pixels. Homogeneity of the pixels may be determined by color and texture features. We have used color feature of the diseased area in order to partition the skin cancer regions. Our proposed method uses histogram and clustering algorithms for segmenting the skin cancer regions.

\subsubsection{Histogram Calculation}
Skin lesions have distinct color which can be differentiated from normal skin. So we have used color feature for segmenting the lesions. Using the ground truth of the training images histograms \cite{gonzalez2009digital} were constructed to find the lower and upper boundary of the pixel color of skin lesions. Three different threshold values are obtained for three types of cancer cells. These obtained threshold values are used to filter the clusters obtained in the next step.
\subsubsection{K-means Clustering}
One of the simplest method to group similar type of data points is k-means clustering. We have used k-means clustering algorithm to group color pixels which have similar color intensive values. K-means clustering algorithm requires number of cluster k apriori. To cluster skin lesions we have applied k-means clustering algorithm to cluster the input images into 5 different clusters. The number of cluster is fixed as 5 since the image contains\cite{Jain2010651}.

\begin{enumerate}
	\item interior skin lesions
	\item boundary of skin lesions
	\item background
	\item noise like air bubbles, rulers and hair
	\item background of disk like structures were the skin samples are placed 	
\end{enumerate}
From the obtained k clusters, the cluster with pixel color which is similar to skin lesion color are retained. 
\subsubsection{Segmentation using Flood Fill}
From the retained clusters the centroid location of the cluster is extracted. This centroid location is set as the seed point to flood fill algorithm to segment the diseased area. Using the cluster centroid as seed point for flood algorithm we can obtain only the middle portion of the skin lesions. So we have also chosen another seed point which is 10 pixel away from the boundary of the obtained cluster. As a result  we have obtained 2 regions of skin lesions, the first region is the middle portion which is the dense diseased skin lesion region and second region is the boundary portion of the obtained first region which is slightly dense diseased area. Finally  both the regions are partitioned together as skin lesion applying the connected component labeling algorithm.

\section{Implementation}
\begin{figure}
\centering
\includegraphics[width=3.5in]{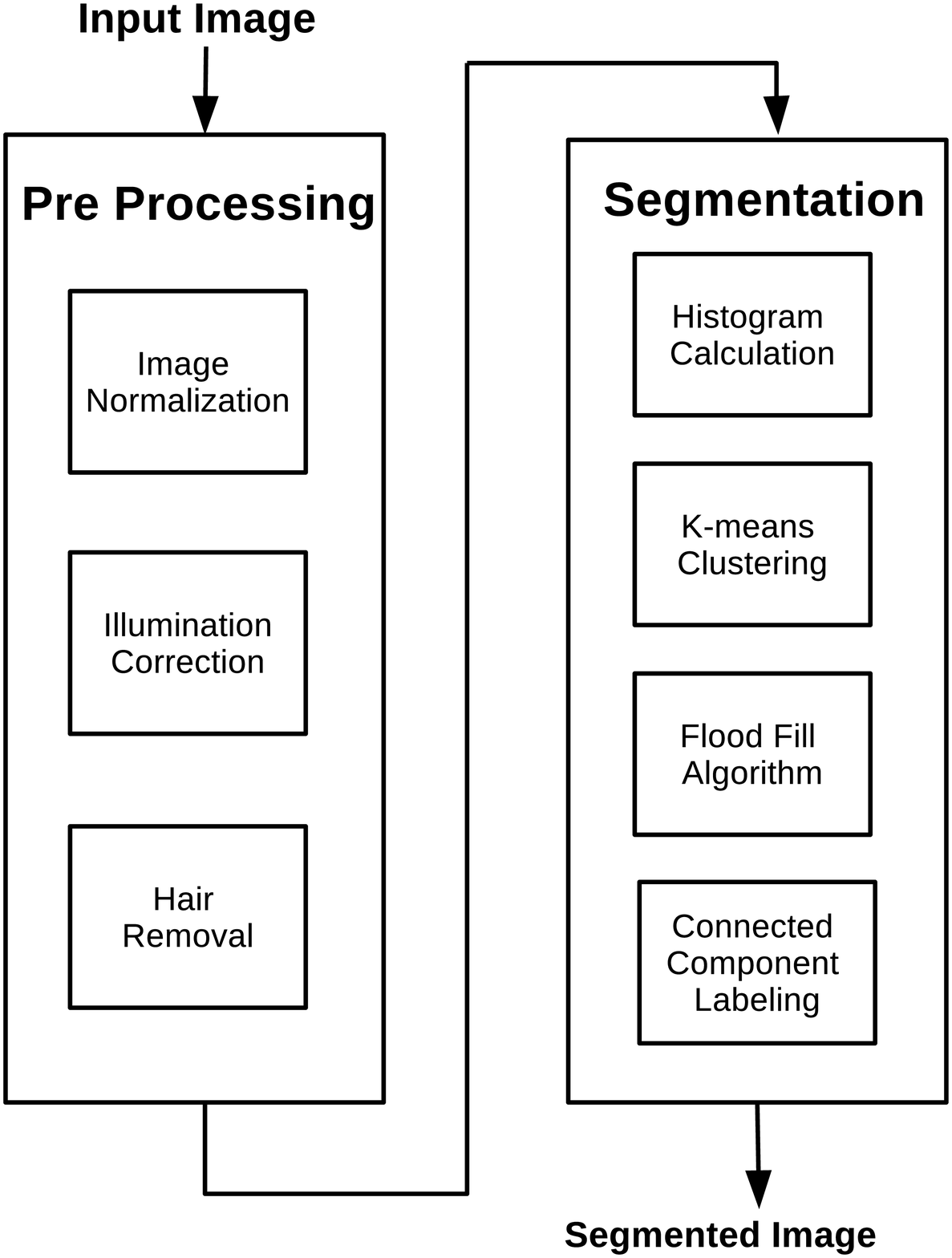}
\caption{ Block Diagram of the Proposed Skin Lesion Segmentation}
\label{fig1}
\end{figure}

The proposed method was implemented using OpenCV. The work flow of the proposed method is depicted in Figure \ref{fig1}.

\section{Results and Discussion}
\subsection{Data Set}
The data set was downloaded from the ISIC 2017: Skin Lesion Analysis Towards Melanoma Detection challenge website.   The training images include 2000 dermoscopic lesion images in JPEG format paired with ground truth information. The ground truth file for each image is the binary mask image with black pixel representing background and white pixel representing  lesion region. The data set also includes optional validation set with 150 images and test set with 600 images. The results of the validation set are evaluated immediately to get some feedback on the performance of the proposed algorithm. 
\subsection{Result}
Our proposed algorithm produced the overall score of 0.548 for validation set. For the validation set we have obtained high sensitivity value and overall accuracy is also high. Only the jaccard index value was less for our proposed method and we are trying to improve this score.   

\section{ Conclusion}
In this research work we have developed  an automatic system for segmenting the skin lesions in the dermoscopic images. The proposed automatic segmentation of skin lesions helps the clinicians to determine the exact location of the lesions for further diagnosis. To segment skin lesions we have first prepossessed the images to correct the illumination variation and to remove artifacts like bubbles and hair. After preprocessing, skin lesions are segmented using semi-supervised algorithm. In the semi-supervised learning technique first the color of skin lesion is determined by constructing histogram using training images. K-means clustering algorithm is applied to images in order to group the pixels of same color. From the obtained clusters the pixels with the skin lesion color alone is retained and then skin lesions are segmented using flood fill algorithm and connected component labeling. For the validation set we have obtained the overall score of 0.548.

\bibliographystyle{IEEEtran}
\bibliography{skin_lesion}

\end{document}